\documentclass[10pt,twocolumn,letterpaper]{article}

\usepackage[pagenumbers]{cvpr} 

\usepackage{graphicx}
\usepackage{amsmath}
\usepackage{amssymb}
\usepackage{booktabs}
\usepackage{multirow}
\usepackage{adjustbox}
\usepackage{enumitem}
\usepackage[accsupp]{axessibility}
\usepackage[pagebackref,breaklinks,colorlinks]{hyperref}

\usepackage[capitalize]{cleveref}
\crefname{section}{Sec.}{Secs.}
\Crefname{section}{Section}{Sections}
\Crefname{table}{Table}{Tables}
\crefname{table}{Tab.}{Tabs.}

\def\cvprPaperID{3963} 
\def\confName{CVPR}
\def\confYear{2022}

\DeclareUnicodeCharacter{0301}{}
\begin{document}

\title{A Text Attention Network for\\ Spatial Deformation Robust Scene Text Image Super-resolution}

\author{Jianqi Ma\textsuperscript{1}, Zhetong Liang\textsuperscript{2}, Lei Zhang\textsuperscript{1}\\
\textsuperscript{1}The Hong Kong Polytechnic University; \textsuperscript{2}OPPO Research\\
{\tt\small \{csjma, cslzhang\}@comp.polyu.edu.hk,}
{\tt\small zhetongliang@163.com}
}
\maketitle


\begin{abstract}
   Scene text image super-resolution aims to increase the resolution and readability of the text in low-resolution images. Though significant improvement has been achieved by deep convolutional neural networks (CNNs), it remains difficult to reconstruct high-resolution images for spatially deformed texts, especially rotated and curve-shaped ones. This is because the current CNN-based methods adopt locality-based operations, which are not effective to deal with the variation caused by deformations. In this paper, we propose a CNN based Text ATTention network (TATT) to address this problem. The semantics of the text are firstly extracted by a text recognition module as text prior information. Then we design a novel transformer-based module, which leverages global attention mechanism, to exert the semantic guidance of text prior to the text reconstruction process. In addition, we propose a text structure consistency loss to refine the visual appearance by imposing structural consistency on the reconstructions of regular and deformed texts. Experiments on the benchmark TextZoom dataset show that the proposed TATT not only achieves state-of-the-art performance in terms of PSNR/SSIM metrics, but also significantly improves the recognition accuracy in the downstream text recognition task, particularly for text instances with multi-orientation and curved shapes. Code is available at \href{https://github.com/mjq11302010044/TATT}{https://github.com/mjq11302010044/TATT}.
   
\end{abstract}

\section{Introduction}
\label{sec:intro}


The text in an image is an important source of information in our daily life, which can be extracted and interpreted for different purposes. However, scene text images often encounter various quality degradation during the imaging process, resulting in low resolution and blurry structures. This problem significantly impairs the performance of the downstream high-level recognition tasks, including scene text detection~\cite{zhou2017east,ma2018arbitrary}, optical character recognition (OCR) and scene text recognition~\cite{shi2016end,shi2018aster,luo2019moran}. Thus, it is necessary to increase the resolution as well as enhance the visual quality of scene text images.

\begin{figure}[t]
  \centering
  \includegraphics[width=\linewidth]{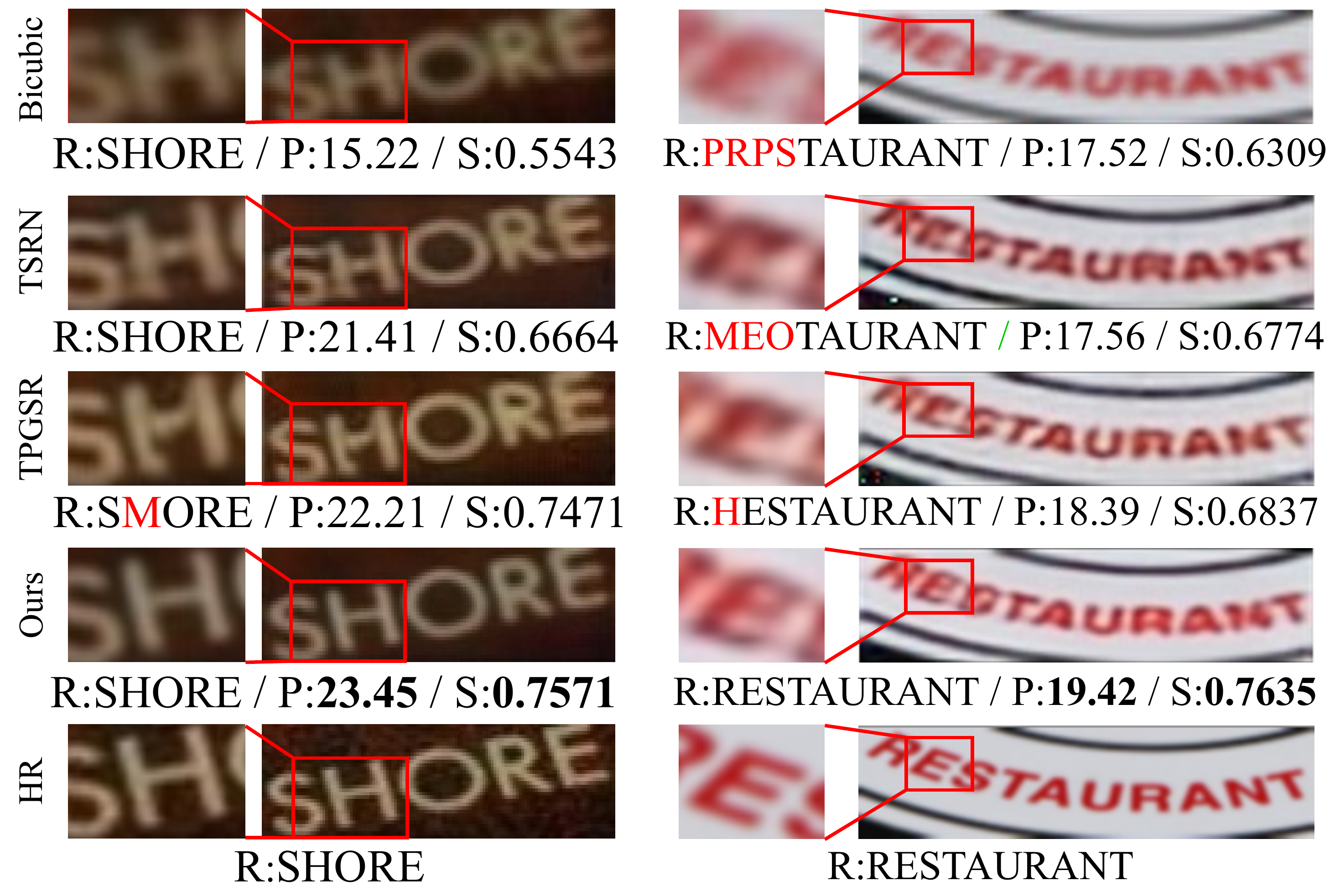}
  \caption{SR recovery of different models on rotated and curve-shaped text images. `R', `P' and `S' stand for recognition, PSNR and SSIM results. Characters in red are missing or wrong.}
  \label{fig:SR_recovery}
  \vspace{-0.5cm}
\end{figure}

In the past few years, many scene text image super-resolution (STISR) methods have been developed to improve the image quality of text images, with notable progress obtained by deep-learning-based methods~\cite{dong2015boosting ,zhang2017cnn,wang2019textsr,wang2020scene,chen2021scene}. By using a dataset of degraded and original text image pairs, a deep convolutional neural network (CNN) can be trained to super-resolve the text image. With strong expressive capability, CNNs can learn various priors from data and demonstrate much strong performance. A recent advance is the TPGSR model~\cite{ma2021text}, where the semantics of the text are firstly recognized as prior information and then used to guide the text reconstruction process. With the high-level prior information, TPGSR can restore the semantically correct text image with compelling visual quality.

Despite the great progress, many CNN-based methods still have difficulty in dealing with spatially-deformed text images, including those with rotation and curved shape. Two examples are shown in Fig.~\ref{fig:SR_recovery}, where the text in the left image has rotation and the right one has a curved shape. One can see that the current representative methods, including TSRN~\cite{wang2020scene} and TPGSR~\cite{ma2021text}, produce blurry texts with semantically incorrect characters. This is because the architectures in current works mainly employ locality-based operations like convolution, which are not effective in capturing the large position variation caused by the deformations. In particular, the TPGSR model adopts a simplistic approach to utilize the text prior: it merely merges text prior with image feature by convolutions. This arrangement can only let the text prior interact with the image feature within a small local range, which limits the effect of text prior on the text reconstruction process. Based on the this observation, some globality-based operations (\eg, attention) should be employed to capture long range correlation in the text image for better STISR performance.   

In this paper, we propose a novel architecture, termed Text ATTention network (TATT), for spatial deformation robust text super resolution. Similar to TPGSR, we first employ a text recognition module to recognize the character semantics as text prior (TP). Then we design a transformer-based module termed TP Interpreter to enforce global interaction between the text prior and the image feature. Specifically, the TP Interpreter operates cross attention between the text prior and the image feature to capture long-range correlation between them. The image feature can then receive rich semantic guidance in spite of the spatial deformation, leading to improved text reconstruction. To further refine the text appearance under spatial deformation, we design a text structure consistency loss, which measures the structural distance between the regular and deformed texts. As can be seen in Fig.~\ref{fig:SR_recovery}, the characters recovered by our method show better visual quality with correct semantics. 

Overall, our contributions can be summarized as follows:

\begin{itemize}[topsep=2pt]
\setlength{\topsep}{0pt}
\setlength{\itemsep}{0pt}
\setlength{\parsep}{0pt}
\setlength{\parskip}{0pt}

\item We propose a novel method to align the text prior with the spatially-deformed text image for better SR recovery by using CNN and Transformer.

\item We propose a text structure consistency loss to enhance the robustness of text structure recovery from spatially-deformed low-resolution text images.

\item Our proposed model not only achieves state-of-the-art performance on the TextZoom dataset in various evaluation metrics, but also exhibits outstanding generalization performance in recovering orientation-distorted and curve-shaped low-resolution text images. 
\end{itemize}

\section{Related Works}

\subsection{Single Image Super Resolution}

Single image super resolution (SISR) aims at recovering a high-resolution (HR) image from a given low-resolution (LR) input image. The traditional methods design hand-crafted image priors for this task, including statistical prior~\cite{GunturkAM04}, self-similarity prior~\cite{MairalBPSZ09} and sparsity prior~\cite{YangWHM10}. The recent deep-learning-based methods train convolutional neural networks (CNNs) to address the SISR task and achieve leading performance. The seminal work SRCNN~\cite{dong2015image} adopts a three-layer CNN to learn the SR recovery. Later on, more complex CNN architectures have been developed to upgrade the SISR performance, \eg, residual block~\cite{lim2017enhanced}, Laplacian pyramid~\cite{lai2017deep}, dense connections~\cite{zhang2018residual} and channel attention mechanism~\cite{zhang2018image}. Recently, generative adversarial networks have been employed in SISR to achieve photo-realistic results~\cite{ledig2017photo,wang2018recovering,chen2018fsrnet}.

\subsection{Scene Text Image Super Resolution (STISR)}

Different from the general purposed SISR that works on natural scene images, STISR focuses on scene text images. It aims to not only increase the resolution of text image, but also reconstruct semantically correct texts that can benefit the down-stream recognition task. The early methods directly adopt the CNN architectures from SISR for the task of STISR. In~\cite{dong2015boosting}, Dong \etal~extended SRCNN~\cite{dong2015image} to text images, and obtained the best performance in ICDAR 2015 competition~\cite{peyrard2015icdar2015}. PlugNet~\cite{mouplugnet} adopts a pluggable super-resolution unit to deal with LR images in feature domain. TextSR~\cite{wang2019textsr} utilizes the text perceptual loss to generate the desired HR images to benefit the text recognition. 

To address the problem of STISR on real-world scenes, Wang~\etal \cite{wang2020scene}~built a real-world STISR image dataset, namely the TextZoom, where the LR and HR text image pairs were cropped from real-world SISR datasets~\cite{zhang2019zoom,cai2019toward}. They also proposed TSRN~\cite{wang2020scene} to use the sequential residual block to exploit the semantic information in internal features. SCGAN~\cite{xu2017learning} employs a multi-class GAN loss to supervise the STISR model for more perceptual-friendly face and text images. Further, Quan \etal~\cite{quan2020collaborative} proposed a cascading model for recovering blurry text images in high-frequency domain and image domain collaboratively. Chen \etal~\cite{chen2021scene} and Zhao~\etal~\cite{zhao2021scene} enhanced the network block structures to improve the STISR performance by self-attending the image features and attending channels.

\begin{figure*}[t]
  \centering
  \includegraphics[width=0.9\linewidth]{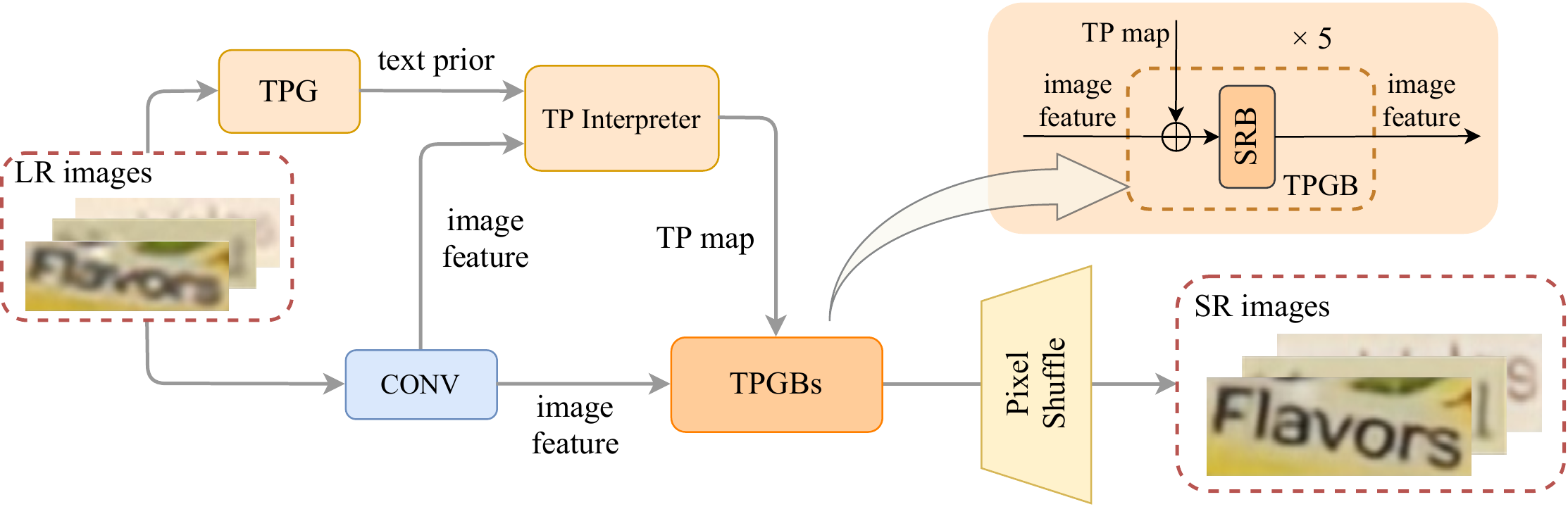}
  \caption{Architecture of our proposed TATT network for STISR. TPGB, TPG and SRB are short for text prior guided blocks, TP Generator and Sequential-Recurrent Blocks, respectively, while $\oplus$ means the element-wise addition.}
  \label{fig:TP-Framework}
  \vspace{-0.3cm}
\end{figure*}

\subsection{Scene Text Recognition}

Scene text recognition aims to extract text content from the input images. 
Some early approaches tend to recognize each character first and then interpret the whole word\cite{jaderberg2014deep,he2015reading}, while some others regard the text image as a whole and performing word-level classification~\cite{jaderberg2016reading}. Considering text recognition as an image-to-sequence problem, CRNN~\cite{shi2016end} extracts image features and uses the recurrent neural networks to model the semantic information. It is trained with CTC~\cite{graves2006connectionist} loss to align the predicted sequence and the target sequence. 
Recently, attention-based methods achieve a great progress due to the robustness in extracting text against shape variance of text images~\cite{cheng2017focusing,cheng2018aon}. Despite the great performance achieved by the recent methods, it is still difficult to recognize the text in low-resolution images. Therefore, we aim to solve the problem of high-resolution text image restoration for better recognition in this paper.

\section{Methodology}


\begin{figure*}[t]
  \centering
  \includegraphics[width=0.85\linewidth]{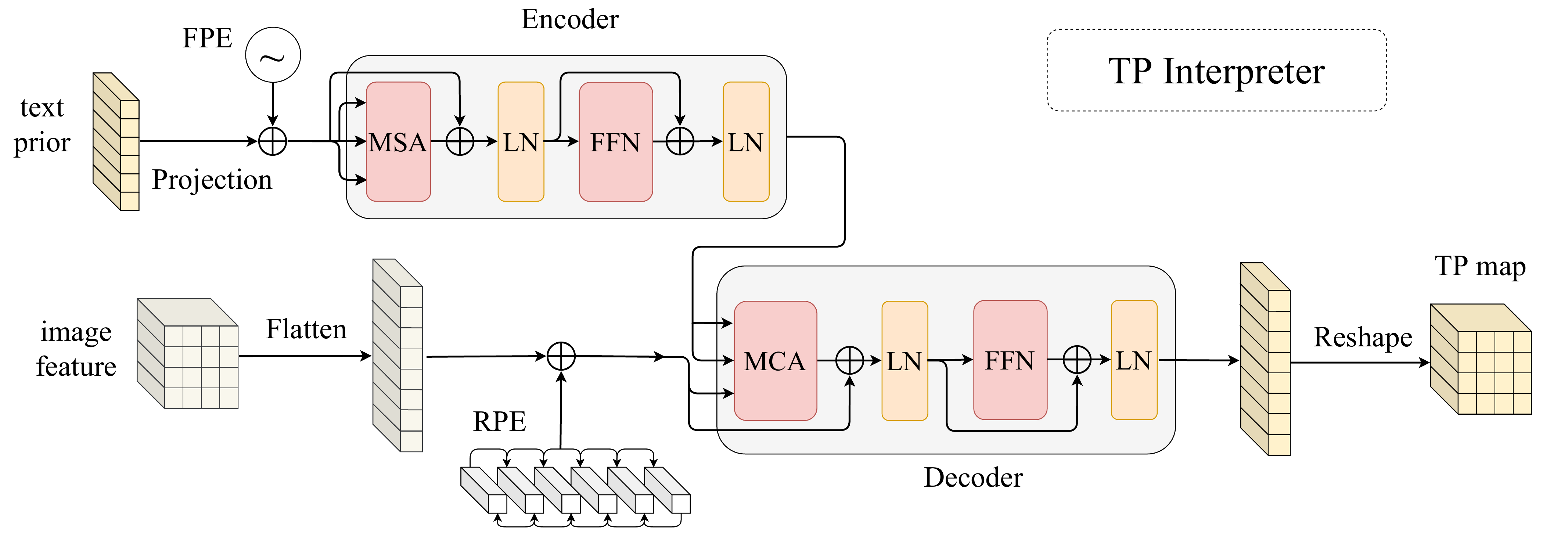}
  \caption{Architecture of TP Interpreter. `MSA', `LN', `MCA' and `FFN' namely mean the Multi-head Self-Attention, Layer-Norm, Multi-head Cross-Attention and Feed-Forward Network layers, while `FPE' and `RPE' refer to the Fixed Positional Encoding and Recurrent Positional Encoding. While $\oplus$ means the element-wise addition.}
  \label{fig:TP-Interpreter}
  \vspace{-0.3cm}
\end{figure*}



\subsection{Overall Architecture}

The pipeline of our TATT network is shown in Fig.~\ref{fig:TP-Framework}. It takes low-resolution (LR) text images $Y \in \mathbb{R}^{h \times w \times 3}$ as input, which is processed in the following two paths. In the first path, the input images are sent into a TP Generator (TPG) to predict the recognition probability sequence as text prior $f_p$ (similar to ~\cite{ma2021text}). This process can be denoted as $f_P = \textit{TPG}(Y)$. $f_P \in \mathbb{R}^{l \times |\mathcal{A}|}$ is an $l$-length sequence composed of categorical probability vectors with size $|\mathcal{A}|$. $\mathcal{A}$ denotes the character set which is composed of `0' to `9', `a' to `z' and a blank class~($37$ in total). The second path extracts image features $f_I \in \mathbb{R}^{h \times w \times c}$ from the input LR image $Y$ by a $9 \times 9$ convolution layer (we denote this process as $f_I = \textit{Conv}(Y)$).

Then, the text prior $f_P$ and the image feature $f_I$ are passed to the TP Interpreter $\textit{TPI}(\cdot)$ to calculate a TP map $f_{TM} \in \mathbb{R}^{h \times w \times c}$, which is denoted as $f_{TM} = \textit{TPI}(f_P, f_I)$. The TP Interpreter computes the correlation between the text prior $f_P$ and image feature $f_I$, and assigns the semantic guidance in $f_P$ to the corresponding location in the spatial domain to guide the final SR text recovery. The resultant TP map $f_{TM}$ is a modulating map which can be use to enhance the semantics-specific part of the image feature . 


Finally, the TP map $f_{TM}$ and the image feature $f_I$ are passed into a reconstruction module. This module includes $5$ Text-Prior Guided Blocks (TPGBs) that progressively fuse $f_{TM}$ and $f_I$, and a final Pixel-Shuffle layer to increase the resolution. Each of the 5 TPGBs firstly merges $f_{TM}$ and $f_I$ by element-wise addition, followed by a Sequential-Recurrent Block~(SRB)~\cite{wang2020scene} to reconstruct the high-resolution image feature. The output of this module is the super-resolved (SR) text image.

\subsection{TP Interpreter}

In the proposed architecture, the crucial part lies in the design of TP Interpreter (TPI). The TP Interpreter aims to interpret the text prior $f_P$ to the image feature $f_I$ so that the influence of the semantics guidance can be exerted to the correlated spatial position in the image feature domain. One intuitive idea is to enlarge $f_P$ to the shape of $f_I$ and then merge them by convolution. Since the convolution operation has a small effective range, the semantics of $f_P$ cannot be assigned to the distant spatial location in $f_I$, especially in the case of spatially-deformed text. Thus, we turn to design a Transformer-based TP Interpreter with attention mechanism to enforce global correlation between text prior $f_P$ and the image feature $f_I$.



As shown in Fig.~\ref{fig:TP-Interpreter}, the proposed TP Interpreter consists of an Encoder part and a Decoder part. The Encoder encodes the text prior $f_P$ by performing correlation between the semantics of each character in $f_P$ and outputs the context-enhanced feature $f_E$. The decoder performs cross attention between $f_E$ and $f_I$ to interpret the semantic information to the image feature. 

\textbf{Encoder.} 
The Encoder takes the text prior $f_P$ as input and project it to $C$ channels to match the image feature channel. Since the input text prior is processed in parallel in the encoder, the model is not aware of the semantic order in TP. We thus encode the position by adding the Fixed Positional Encoding (FPE) to $f_P$ in an element-wise manner before feeding it into the encoder. Note that we adopt Sinusoidal Positional Encoding~\cite{vaswani2017attention} as our FPE in this paper.
After encoding the position, the text prior is passed into the encoder module. The encoder has a Multi-head Self Attention (MSA) layer and a FeedForward Network (FFN) layer~\cite{vaswani2017attention}. Skip connection is deployed between the current layer and the previous layer to enable residual learning. The MSA layer performs global correlation between the semantic elements in text prior $f_P$, resulting in a contextually enhanced TP feature $f_E \in \mathbb{R}^{l \times c}$ for later computation. Due to the space limit, the description of MSA and FFN is omitted. One can refer to \cite{vaswani2017attention} for details.


\textbf{Decoder.} The decoder module accepts the output from the encoder module $f_E$ and image feature $f_I$ to perform global cross attention. Similar to the setting in encoder, we firstly add a position encoding to $f_I$ to incorporate position information. We design a recurrent positional encoding (RPE) to better encode the bias contained in sequential dependency of image feature in horizontal direction, and better help the model look up the text semantic features in the subsequent cross attention~\cite{shiv2019novel,liutkus2021relative}.
In RPE, we maintain the learnable parameter with the same shape as image feature and encode the sequential dependency in horizontal direction to help the model better learn the neighboring context.
See \textbf{supplementary file} for more details. 

The position-encoded image feature, denoted by $f_I^{'}$, and the encoder output $f_E$ are then delivered to the decoder module for correlation computation. 
We process the two inputs with a Multi-head Cross Attention ($\textit{MCA}$) layer, which performs cross attention operation between $f_E$ and $f_I^{'}$. Firstly, the features of $f_E$ and $f_I^{'}$ are divided into $n$ subgroups in the channel dimension. Then a cross attention operation $\textit{CA}_i$ is performed on the $i$-th group of $f_E$ and $f_I^{'}$:
\begin{equation}
\textit{CA}_i(f_{Ei}, f_{Ii}^{'}) = \textit{SM}(\frac{(f_{Ii}^{'}W_i^{\alpha})  (f_{Ei}W_i^{\beta})^{T}}{\sqrt{d_k}})(f_{Ei}W_i^{\gamma})
\label{equCA}
\end{equation}

\noindent where $f_{Ei}\in\mathbb{R}^{l\times\frac{c}{n}}$ and $f_{Ii}^{'}\in\mathbb{R}^{hw\times\frac{c}{n}}$ denote the $i$-th group of $f_E$ and $f_I^{'}$, respectively. $W_i^{\alpha} \in \mathbb{R}^{\frac{c}{n} \times d_k}$, $W_i^{\beta} \in \mathbb{R}^{\frac{c}{n} \times d_k}$ and $ W_i^{\gamma} \in \mathbb{R}^{\frac{c}{n} \times d_k}$ are the parameters of linear projections. $SM$ refers to the Softmax operation. 
We process the results ${\textit{CA}}_i$~($i \in \{0,1,...,n-1\}$) with a channel-wise concatenation $\odot(\cdot)$ and a linear projection $W^o$, described as
\begin{equation}
\textit{MCA} = \odot(\textit{CA}_0, \textit{CA}_1, ..., \textit{CA}_{n-1})W^o
\label{equDecoder}
\end{equation}

\noindent The output of MCA is passed to a FFN for feature refinement, and then reshaped to obtain the TP map $f_{TM}$.

By using the MCA operation, the text prior $f_E$ can effectively interact with the image feature $f_I^{'}$ by correlating every element in semantic domain to the position in spatial domain. Thus, the semantically meaningful regions in the spatial domain are strengthened in the TP map $f_{TM}$, which can be used to modulate the image feature for semantic-specific text reconstruction.

\subsection{Text Structure Consistency Loss}
While the proposed TATT network can attain a good performance, the reconstructed text image still needs some refinement to improve the visual appearance. This is because it is a bit difficult for a CNN model to represent the deformed text features as it does for regular text features, and the reconstructed text image has weaker character structures with relatively low contrast. As a remedy, we simulate deformed text images and design a text structure consistency (TSC) loss to train the proposed TATT network. 

We consider minimizing the distance of three images, \ie, the deformed version of the SR text image $\mathbf{D}\mathcal{F}(Y)$, the SR version of the deformed LR text image $\mathcal{F}(\mathbf{D}Y)$ and the deformed ground truth $\mathbf{D}(X)$, where $\mathbf{D}$ denotes the random deformation\footnote{We consider rotation, shearing and resizing in this paper.}. By increasing the similarity among the three items, we can encourage the CNN model to reduce the performance drop when encountering spatial deformations. The proposed TSC loss firstly measures the structural similarity between the above triplet. For this purpose, we extend the Structure-Similarity Index Measure (SSIM)~\cite{wang2004image} to a triplex SSIM (TSSIM), described as
%
\begin{equation}
\begin{array}{l}
    \begin{split}
        &\textit{TSSIM}(x, y, z) =\\ &\frac{(\mu_{x}\mu_{y} + \mu_{y}\mu_{z} + \mu_{x}\mu_{z} + C_1)(\sigma_{xy} + \sigma_{yz} + \sigma_{xz} + C_2)}{(\mu_{x}^2 + \mu_{y}^2 + \mu_{z}^2 + C_1)(\sigma_{x}^2 + \sigma_{y}^2 + \sigma_{z}^2 + C_2)}
    \end{split}
\end{array}
\label{equ:TRI_SSIM}
\end{equation}

\noindent where $\mu_x$, $\mu_y$, $\mu_z$ and $\sigma_x$, $\sigma_y$, $\sigma_z$ represent the mean and standard deviation of the triplet $x$, $y$ and $z$, respectively. $\sigma_{xy}$, $\sigma_{yz}$ and $\sigma_{xz}$ denote the correlation coefficients between $(x, y)$, $(y, z)$ and $(x, z)$, respectively. $C_1$ and $C_2$ are small constants to avoid instability for dividing values close to zero. The derivation is in the \textbf{supplementary file}.

Lastly, TSC loss $L_{TSC}$ is designed to measure the mutual structure difference among $\mathbf{D}\mathcal{F}(Y)$, $\mathcal{F}(\mathbf{D}Y)$ and $\mathbf{D}X$:
\begin{equation}
\begin{array}{l}
    \begin{split}
&{L}_{TSC}(X, Y;\mathbf{D}) = \\
&1 - \textit{TSSIM}(\mathbf{D}\mathcal{F}(Y), \mathcal{F}(\mathbf{D}Y), \mathbf{D}X)
\end{split}
\end{array}
\label{equ:TSCLoss}
\end{equation}



\subsection{Overall Loss Function}
In the training, the overall loss function includes a super resolution loss $L_{SR}$, a text prior loss $L_{TP}$ and the proposed TSC loss $L_{TSC}$. The SR loss $L_{SR}$ measures the difference between our SR output $\mathcal{F}(Y)$ and the ground-truth HR image $X$. We adopt $L_2$ norm for this computation. The TP loss measures the $L_1$ norm and KL Divergence between the text prior extracted from the LR image and those from the ground truth. Together with TSC loss $L_{TSC}$, the overall loss function is described as follows:
\begin{equation}
{L} = L_{SR} + \alpha  L_{TP} + \beta L_{TSC}
\label{equ_AllLoss}
\end{equation}
where the $\alpha$ and $\beta$ are the balancing parameters.

\section{Experiments}

\subsection{Implementation Details}

TATT is trained and tested on a single RTX 3090 GPU.
We adopt Adam~\cite{kingma2014adam} optimizer to train the model with batch size $64$. The training lasts for $500$ epochs with learning rate $10^{-3}$. The input image of our model is of width $64$ and height $16$, while the output is the $2 \times$ SR result. We set the $\alpha$ and $\beta$ in (\ref{equ_AllLoss}) to $1$ and $0.1$, respectively~(see \textbf{supplementary file} for ablations). The deformation operation $\mathbf{D}$ in $L_{TSC}$ is implemented by applying random rotation in a range of $[-10, 10]$ degree, shearing and aspect ratio in a range of $[0.5, 2.0]$. The head numbers of MSA and MCA layers are both set to $4$ (following the best settings in ~\cite{carion2020end}). The number of image feature channels $c$, $d_k$ in MSA, MCA and FFN calculation are all set to $64$. The model size of TATT is $14.9$M in total. When training, the TPG is initialized with pretrained weights derived from ~\cite{crnnpytorch}, while other parts are randomly initialized. When testing, TATT will occupy 6.5GB of GPU memory with batch size $50$.


\subsection{Datasets}

\textbf{TextZoom.} TextZoom~\cite{wang2020scene} has $21,740$ LR-HR text image pairs collected by changing the focal length of the camera in real-world scenarios, in which $17,367$ samples are used for training. The rest samples are divided into three subsets, based on the camera focal length, for testing , namely easy ($1,619$ samples), medium ($1,411$ samples) and hard ($1,343$ samples). Text label is provided in TextZoom.

\textbf{Scene Text Recognition Datasets.} Besides experiments conducted in TextZoom, we also adopt ICDAR2015~\cite{karatzas2015icdar}, CUTE80~\cite{risnumawan2014robust} and SVTP~\cite{phan2013recognizing} to evaluate the robustness of our model in recovering spatially-deformed LR text images. ICDAR2015 has $2,077$ scene text images for testing. Most text images suffer from both low quality and perspective-distortion, making the recognition extremely challenging. CUTE80 is also collected in the wild. The test set has $288$ samples in total. Samples in SVTP are mostly curve-shaped text. The total size of the test set is $649$. Besides evaluating our model on the original samples, we further degrade the image quality to test the model generalization against unpredicted bad conditions. 

\subsection{Ablation Studies}
\label{sec:AblationStudy}
In this section, we investigate the impact of TP Interpreter, the TSC loss function and the effectiveness of positional encoding. All evaluations in this section are performed on the real-world STISR dataset TextZoom~\cite{wang2020scene}. The text recognition is peformed by CRNN~\cite{shi2016end}.


\textbf{Impact of TP Interpreter in SR recovery.}
Since our TP Interpreter aims at providing better alignment between TP and the image feature and use text semantics to guide SR recovery, we compare it with other guiding strategies, \eg, first upsampling the TP to match the image feature with deconvolution layers~\cite{ma2021text} or pixel shuffle to align text prior to image feature, and then fusing them to perform guidance with element-wise addition or SFT layers~\cite{wang2018recovering}\footnote{The SFT layer merges the semantics of image feature with channel-wise affine transformation.}. The results are shown in Tab.~\ref{table:Fusion_ablation}. One can see that the proposed TP interpreter obtains that highest PSNR/SSIM, which also indicates the best SR performance. 

\begin{figure}[t]
  \centering
  \includegraphics[width=\linewidth]{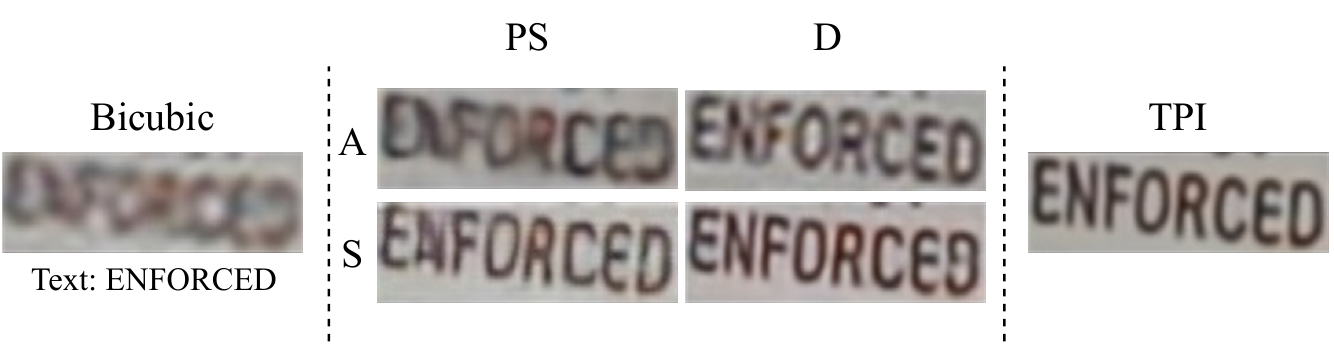}
  \caption{SR recovery by different guiding strategies.}
  \label{fig:TPI_COM}
  \vspace{-0.3cm}
\end{figure}

Referring to the SR text image recognition, one can see that using Pixel-Shuffle and deconvolution strategies provides inferior guidance ($46.2\%$ and $49.8\%$). There is no stable improvement by combining them with the SFT layers ($47.9\%$ and $48.6\%$). This is because none of the competing strategies performs global correlation between the text semantics and the image feature, resulting in inferior semantic guidance for SR recovery. In contrast, our TP Interpreter can obtain a good semantics context and accurate alignment to the text region. It thus strengthens the guidance in image feature and improves the text recognition result to $52.6\%$. This validates that using TP Interpreter is an effective way to utilize TP semantics for SR recovery. Some visual comparisons are shown in Fig.~\ref{fig:TPI_COM}. One can see that the setting with TP interpreter can lead to the highest quality SR text image with correct semantics.

\begin{table}
\small
\centering
\begin{tabular}{l|ccc}
\hline
 Strategy & avg & PSNR & SSIM \\\hline
 w/o TP &  41.4\% & 21.42 & 0.7690\\\hline
  PS + A & 46.2\% & 20.58 & 0.7683\\
  PS + S~\cite{wang2018recovering} & 47.9\% & 20.72 & 0.7560\\
  D~\cite{ma2018arbitrary} + A &  50.6\% & 21.10 & 0.7819\\
  D~\cite{ma2018arbitrary} + S~\cite{wang2018recovering} & 49.6\% & 20.87 & 0.7783 \\
  TPI & \textbf{52.6\%}& \textbf{21.52}& \textbf{0.7930}\\\hline
 
\end{tabular}
\caption{Modules adopted in aligning and guiding the TP sequence to the image feature. D and PS refer to aligning operations Deconvolution and Pixel-Shuffle, respectively. A and S refer to guidance fusion operations by element-wise Addition and SFT Layers~\cite{wang2018recovering}, respectively. TPI is the TP Interpreter.}
\label{table:Fusion_ablation}
\vspace{-0.3cm}
\end{table}%

\begin{figure}[t]
  \centering
  \includegraphics[width=\linewidth]{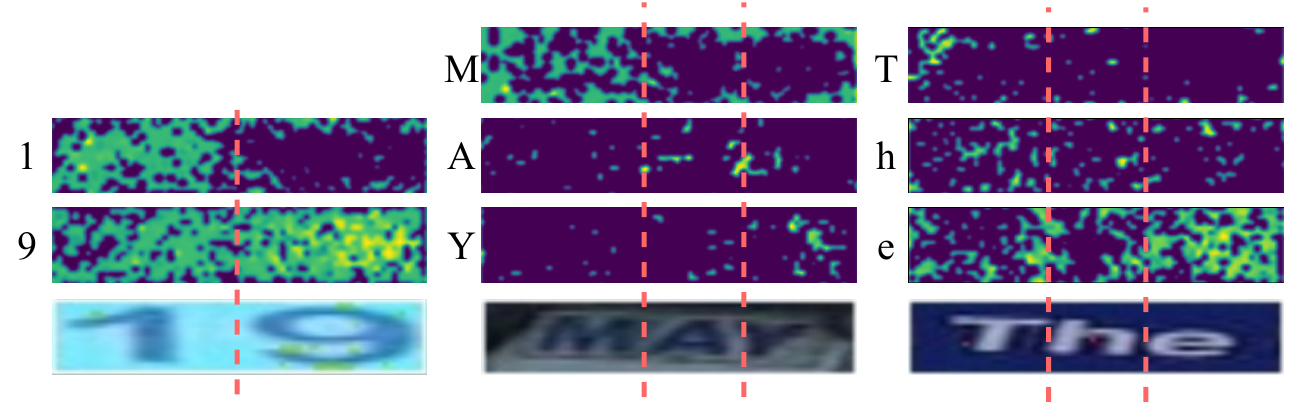}
  \caption{Attention heatmap of the foreground characters.}
  \label{fig:AttentionHeatmap}
  \vspace{-0.3cm}
\end{figure}

To demonstrate how the TP Interpreter provides global context, we visualize the attention heatmap provided by our MCA (the outputs from the $SM$ layer in (\ref{equCA})) in Fig.~\ref{fig:AttentionHeatmap}. One can see that the region of the corresponding foreground character has the highest weight (highlighted). It thus proves that the ability of TP Interpreter in finding semantics in image features. Some other highlighted regions in the neighborhood also demonstrate that the TP Interpreter can be aware of the neighboring context, which can provide better guidance for final SR recovery. 
%


\textbf{Impact of training with TSC loss.} To validate the effectiveness of the TSC loss in refining text structure, we compare the results of $4$ models trained with and without the TSC loss, including non-TP based TSRN~\cite{wang2020scene}, TBSRN~\cite{chen2021scene}, TP based TPGSR~\cite{ma2021text} and TATT. From the results in Tab.~\ref{table:LossImpact_1}, one can see that all models lead to a performance gain ($4.3\%$ for TSRN, $1.3\%$ for TBSRN, $0.8\%$ for TPGSR, and $1.0\%$ for TATT) in SR text recognition when adopting our TSC loss. Notably, though TBSRN~\cite{chen2021scene} is claimed to be robust for multi-oriented text, it can still be improved with our TSC loss, indicating that training with the TSC loss can improve the robustness of reconstructing the character structure against various spatial deformations. 

\textbf{Effectiveness of the RPE.} 
We evaluate the impact of recurrent positional encoding in learning text prior guidance. We deploy different combinations of fixed positional encoding (FPE), learnable positional encoding~\cite{carion2020end} and the proposed recurrent positional encoding (RPE) in the encoder and decoder modules, and compare the corresponding text recognition results on the SR text images. From Tab.~\ref{table:RPE}, we observe that using LPE or FPE in decoder shows limited performance because they are weak in learning the sequential information. By adopting RPE in the decoder, the SR recognition is increased by $1.8\%$, indicating that RPE is beneficial to text sequential semantics learning.


\begin{table}
\small
\centering
\setlength\tabcolsep{4pt}
\begin{tabular}{l|c|cccc}
\hline
 Approach & $L_{\textit{TSC}}$ & easy & medium & hard & avg \\\hline
  \multirow{2}{*}{TSRN~\cite{wang2020scene}} & $\times$ & 52.5\% & 38.2\% & 31.4\% & 41.4\%\\
  & $\checkmark$ & \textbf{58.0\%} & \textbf{43.2\%} & \textbf{33.4\%} & \textbf{45.7\%}\\\hline
  \multirow{2}{*}{TBSRN~\cite{chen2021scene}} & $\times$ & 59.6\% & 47.1\% & 35.3\% & 48.1\%\\
  & $\checkmark$ & \textbf{60.8\%} & \textbf{49.6\%} & \textbf{36.1\%} & \textbf{49.4\%}\\\hline
  \multirow{2}{*}{TPGSR~\cite{ma2021text}} & $\times$ & 61.0\% & \textbf{49.9\%} & 36.7\% & 49.8\%\\
   & $\checkmark$ & \textbf{62.0\%} & 49.8\% & \textbf{37.4\%} & \textbf{50.6\%} \\\hline
  \multirow{2}{*}{ours} & $\times$ & 62.1\% & 52.1\% & 37.8\% & 51.6\%\\
   & $\checkmark$ & \textbf{62.6\%} & \textbf{53.4\%} & \textbf{39.8\%} & \textbf{52.6\%} \\\hline
\end{tabular}
\caption{TextZoom results of models with and without TSC loss.}
\label{table:LossImpact_1}
\end{table}%

\begin{table}
\small
\centering
\begin{tabular}{c|ccc}
\hline
 Approach & Enc & Dec & avg \\\hline
 \multirow{3}{*}{Ours} & FPE & FPE & 50.5\% \\
  & FPE & LPE & 50.8\% \\
  & FPE & RPE & \textbf{52.6\%} \\\hline
\end{tabular}
\caption{SR text image recognition results of different positional encoding ablations on TextZoom. The Enc and Dec refer to the encoder and decoder of the TP Interpreter.}
\label{table:RPE}
\vspace{-0.3cm}
\end{table}%

\begin{table*}
\small
\centering
\begin{tabular}{l|c|cccc|cccc}
\hline
~ & ~ & \multicolumn{4}{c|}{PSNR} & \multicolumn{4}{c}{SSIM}\\\hline
Method & Loss & easy & medium & hard & \textbf{avg} & easy & medium & hard & \textbf{avg} \\\hline
Bicubic & $\times$ & 22.35 & 18.98 & 19.39 & 20.35 & 0.7884 & 0.6254 & 0.6592 & 0.6961 \\\hline
SRCNN~\cite{dong2015image} & $L_2$ & 23.48 & 19.06 & 19.34 & 20.78 & 0.8379 & 0.6323 & 0.6791 & 0.7227 \\
SRResNet~\cite{ledig2017photo} & $L_2$+$L_{\textit{tv}}$+$L_p$  & 24.36 & 18,88 & 19.29 & 21.03 & 0.8681 & 0.6406 & 0.6911 & 0.7403 \\
HAN~\cite{niu2020single} & $L_2$ & 23.30 & 19.02 & 20.16 & 20.95 & 0.8691 & 0.6537 & 0.7387 & 0.7596\\
TSRN~\cite{wang2020scene} & $ L_2$+$L_{\textit{GP}}$ & \textbf{25.07} & 18.86 & 19.71 & 21.42 & 0.8897 & 0.6676 & 0.7302 & 0.7690 \\
TBSRN~\cite{ma2021text} & $L_{\textit{POS}}$+$L_{\textit{CON}}$ & 23.46 & \textbf{19.17} & 19.68 & 20.91 & 0.8729 & 0.6455 & 0.7452 & 0.7603 \\
PCAN~\cite{zhao2021scene} & $L_2$+$L_{\textit{EG}}$ & 24.57 & 19.14 & 20.26 & 21.49 & 0.8830 & 0.6781 & 0.7475 & 0.7752\\
TPGSR~\cite{ma2021text} & $L_2$+$L_{\textit{TP}}$ & 23.73 & 18.68 & 20.06 & 20.97 & 0.8805 & 0.6738 & 0.7440 & 0.7719 \\
TPGSR-3~\cite{ma2021text} & $L_2$+$L_{\textit{TP}}$ & 24.35 & 18.73 & 19.93 & 21.18 & 0.8860 & 0.6784 & 0.7507 & 0.7774 \\
TATT & $L_2$+$L_{\textit{TP}}$+$L_{\textit{TSC}}$ &  24.72 & 19.02 & \textbf{20.31} & \textbf{21.52} & \textbf{0.9006} & \textbf{0.6911} & \textbf{0.7703} & \textbf{0.7930} \\\hline

\end{tabular}
\caption{PSNR/SSIM indices for competing SISR and STISR methods. `-3' means multi-stage settings in~\cite{ma2021text}.}
\label{table:PSNRSSIM}
\end{table*}%

\begin{table*}
\centering
\setlength\tabcolsep{4pt}
\scalebox{0.98}{
\footnotesize
\begin{tabular}{l|c|cccc|cccc|cccc}
\hline
~ & ~ & \multicolumn{4}{c|}{ASTER~\cite{shi2018aster}} & \multicolumn{4}{c|}{MORAN~\cite{luo2019moran}} & \multicolumn{4}{c}{CRNN~\cite{shi2016end}}\\\hline
Method & Loss & easy & medium & hard & \textbf{avg} & easy & medium & hard & \textbf{avg} & easy & medium & hard & \textbf{avg}\\\hline
Bicubic & $\times$ & 64.7\% & 42.4\% & 31.2\% & 47.2\% & 60.6\% & 37.9\% & 30.8\% & 44.1\% & 36.4\% & 21.1\% & 21.1\% & 26.8\%\\\hline
SRCNN~\cite{dong2015image} & $L_2$ & 69.4\% & 43.4\% & 32.2\% & 49.5\% & 63.2\% & 39.0\% & 30.2\% & 45.3\% & 38.7\% & 21.6\% & 20.9\% & 27.7\%\\
SRResNet~\cite{ledig2017photo} & $L_2$+$L_{\textit{tv}}$+$L_p$ & 69.4\% & 47.3\% & 34.3\% & 51.3\% & 60.7\% & 42.9\% & 32.6\% & 46.3\% & 39.7\% & 27.6\% & 22.7\% & 30.6\%\\
HAN~\cite{niu2020single} & $L_2$ & 71.1\% & 52.8\% & 39.0\% & 55.3\% & 67.4\% & 48.5\% & 35.4\% & 51.5\% & 51.6\% & 35.8\% & 29.0\% & 39.6\%\\
TSRN~\cite{wang2020scene} & $ L_2$+$L_{\textit{GP}}$ & 75.1\% & 56.3\% & 40.1\% & 58.3\% & 70.1\% & 53.3\% & 37.9\% & 54.8\% & 52.5\% & 38.2\% & 31.4\% & 41.4\%\\
TBSRN~\cite{ma2021text} & $L_{\textit{POS}}$+$L_{\textit{CON}}$ & 75.7\% & 59.9\% & 41.6\% & 60.0\% & 74.1\% & 57.0\% & 40.8\% & 58.4\% & 59.6\% & 47.1\% & 35.3\% & 48.1\%\\
PCAN~\cite{zhao2021scene} & $L_2$+$L_{\textit{EG}}$ & 77.5\% & 60.7\% & 43.1\% & 61.5\% & 73.7\% & 57.6\% & 41.0\% & 58.5\% & 59.6\% & 45.4\% & 34.8\% & 47.4\%\\
TPGSR~\cite{ma2021text} & $L_2$+$L_{\textit{TP}}$ & 77.0\% & 60.9\% & 42.4\% & 60.9\% & 72.2\% & 57.8\% & 41.3\% & 57.8\% & 61.0\% & 49.9\% & 36.7\% & 49.8\%\\
TPGSR-3~\cite{ma2021text} & $L_2$+$L_{\textit{TP}}$ & \textbf{78.9\%} & 62.7\% & 44.5\% & 62.8\% & \textbf{74.9\%} & \textbf{60.5\%} & \textbf{44.1\%} & \textbf{60.5\%} & \textbf{63.1\%} & 52.0\% & 38.6\% & 51.8\%\\
TATT & $L_2$+$L_{\textit{TP}}$+$L_{\textit{TSC}}$ & \textbf{78.9\%} & \textbf{63.4\%} & \textbf{45.4\%} & \textbf{63.6\%} & 72.5\% & 60.2\% & 43.1\% & 59.5\% & 62.6\% & \textbf{53.4\%} & \textbf{39.8\%} & \textbf{52.6\%} \\\hline
HR & - &  94.2\% & 87.7\% & 76.2\%  & 86.6\% &  91.2\% & 85.3\%  & 74.2\%  &  84.1\% & 76.4\%  & 75.1\%  & 64.6\%  &  72.4\%\\\hline
\end{tabular}}
\caption{SR text recognition for competing SISR and STISR methods. `-3' means multi-stage settings in~\cite{ma2021text}.}
\label{table:SR_text_recognition}
\vspace{-0.3cm}
\end{table*}%

\begin{table}
\centering
\footnotesize
\begin{tabular}{l|ccc|cc}
\hline
   Method & AS~\cite{shi2018aster} & MO~\cite{luo2019moran} & CR~\cite{shi2016end} & PSNR & SSIM \\\hline
   Bicubic~\cite{wang2020scene} &  36.1\% & 32.2\% & 19.5\% & 19.68 & 0.6658\\
   TSRN~\cite{wang2020scene} &  46.6\%& 43.8\% & 35.2\% &19.70 & 0.7157\\
   TBSRN~\cite{chen2021scene} &  48.5\%& 45.1\% & 37.3\% & 19.10 & 0.7066 \\
   TPGSR~\cite{ma2018arbitrary} & 46.6\% & 45.3\% & 40.2\% &19.79 & 0.7293\\
   Ours & \textbf{51.7\%} & \textbf{47.3\%} & \textbf{43.8\%} & \textbf{20.20} & \textbf{ 0.7535}\\\hline
   HR &  80.8\% & 75.7\% & 68.8\% & - & -\\\hline
\end{tabular}
\caption{Evaluation of competitive STISR models on spatially-deformed samples picked in TextZoom in terms of recognition, PSNR and SSIM. `AS', `MO' and `CR' refer to ASTER~\cite{shi2018aster}, MORAN~\cite{luo2019moran} and CRNN~\cite{shi2016end}, respectively.}
\label{table:DistortedTextZoom}
\vspace{-0.3cm}
\end{table}%

\begin{figure*}[t]
  \centering
  \includegraphics[width=\linewidth]{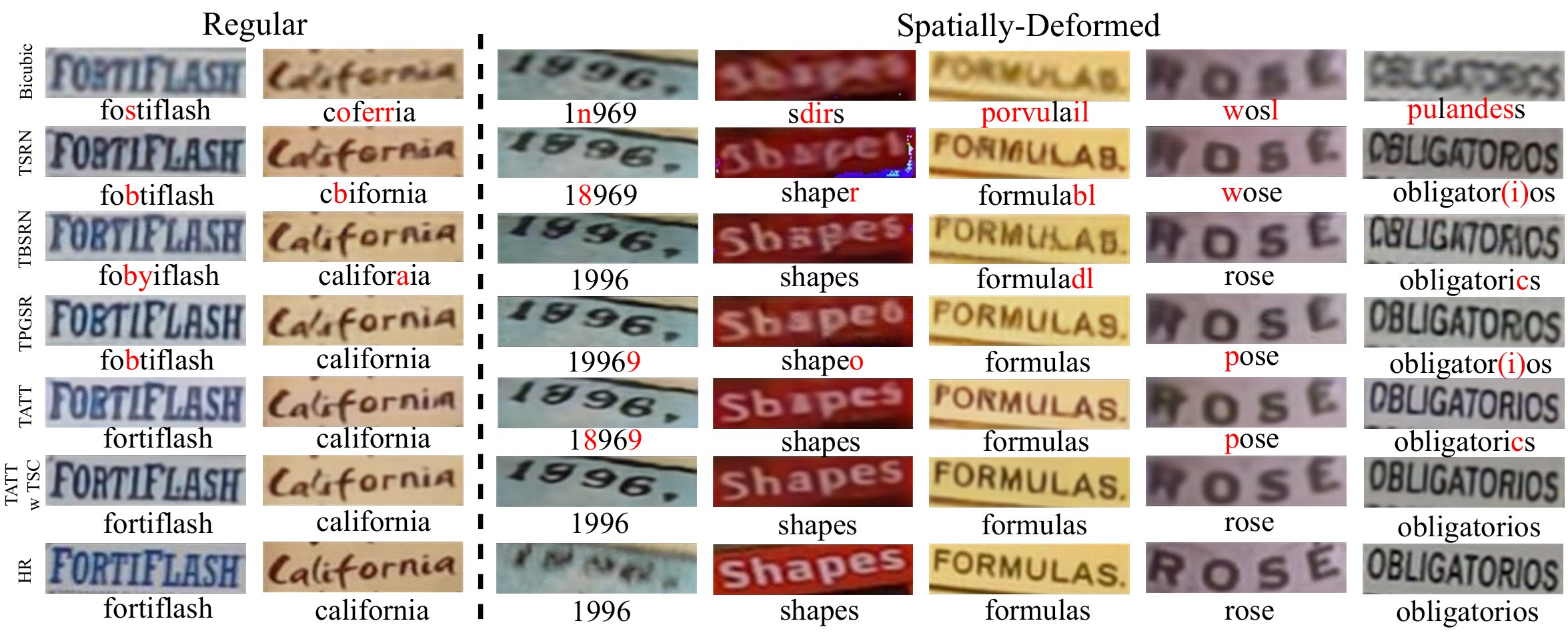}
  \caption{Visualization of regular and spatially-deformed samples from TextZoom recovered by state-of-the-art STISR models and the SR text recognition results. Characters in red are missing or wrong. `w TSC' means that the model is trained with our TSC loss.}
  \label{fig:TextZoomVis}
  \vspace{-0.3cm}
\end{figure*}


\begin{table}
\footnotesize
\centering
\begin{tabular}{c|l|ccc}
\hline
   & super-resolver & AS~\cite{shi2018aster} & MO~\cite{luo2019moran} & CR~\cite{shi2016end}\\\hline
   \multirow{5}{*}{O} & Bicubic & 38.1\% & 29.1\% & 18.1\% \\
    & TSRN~\cite{wang2020scene}  & 41.5\% & 33.8\% & 26.6\% \\
    & TBSRN~\cite{chen2021scene}  & 46.8\% & 45.3\% & 38.3\% \\
    & TPGSR~\cite{ma2021text} & 53.1\% & 52.3\% & 42.5\%  \\
    & Ours & \textbf{53.4\%} & \textbf{59.1\%} & \textbf{47.2\%} \\\hline
    \multirow{5}{*}{CO} & Bicubic&33.2\%&28.1\%&23.6\%\\
    &TSRN~\cite{wang2020scene}&46.4\%&42.1\%&29.1\%\\
    &TBSRN~\cite{chen2021scene}&45.5\%&44.7\%&31.9\%\\
    &TPGSR~\cite{ma2021text}&48.3\%&52.8\%&38.3\%\\
    &Ours&\textbf{54.7\%}&\textbf{54.0\%}&\textbf{45.1\%}\\\hline
    \multirow{5}{*}{GN} & Bicubic&29.4\%&25.8\%&7.5\%\\
    &TSRN~\cite{wang2020scene}&31.3\%&27.5\%&11.5\%\\
    &TBSRN~\cite{chen2021scene}&40.2\%&\textbf{33.4\%}&15.8\%\\
    &TPGSR~\cite{ma2021text}&35.7\%&31.7\%&18.1\%\\
    &Ours&\textbf{43.0\%}&\textbf{33.4\%}&\textbf{21.1\%}\\\hline
    \multirow{5}{*}{GB} &Bicubic&27.0\%&22.3\%&5.5\%\\
    &TSRN~\cite{wang2020scene}&39.2\%&35.8\%&20.4\%\\
    &TBSRN~\cite{chen2021scene}&42.6\%&42.8\%&20.8\%\\
    &TPGSR~\cite{ma2021text}&45.9\%&\textbf{43.8\%}&29.6\%\\
    &Ours&\textbf{47.4\%}&\textbf{43.8\%}&\textbf{35.7\%}\\\hline

\end{tabular}
\caption{Impact of using different STISR models as super-resolver against degradation. `O', `CO', `GB' and `GN' refer to original images and image degradation in terms of contrast, Guassian blurring and Gaussian noise. `AS', `MO' and `CR' refer to ASTER~\cite{shi2018aster}, MORAN~\cite{luo2019moran} and CRNN~\cite{shi2016end}, respectively.}
\label{table:SynLR_Ablations}
\vspace{-0.3cm}
\end{table}%

\subsection{Comparison with State-of-the-Arts}

\textbf{Results on TextZoom.} We conduct experiments on the real-world STISR dataset TextZoom~\cite{wang2020scene} to compare the proposed TATT network with state-of-the-art SISR models, including SRCNN~\cite{dong2015image} and SRResNet~\cite{ledig2017photo} and HAN~\cite{niu2020single}, and STISR models, including TSRN~\cite{wang2020scene}, TPGSR~\cite{ma2021text}, PCAN~\cite{zhao2021scene} and TBSRN~\cite{chen2021scene}. For TPGSR, we compare two models of it, \ie, 1-stage and 3-stage (TPGSR-3). The evaluation metrics are SSIM/PSNR and text recognition accuracy. The comparison results are shown in Tab.~\ref{table:PSNRSSIM} and Tab.~\ref{table:SR_text_recognition}. 

One can see that our model trained with $L_{TSC}$ achieves the best PSNR ($21.52$) and SSIM ($0.7930$) overall performance. This verifies the superiority of our method in improving the image quality. 
As for the SR text recognition, our method achieves new state-of-the-art accuracy under all settings by using the text recognition models of ASTER~\cite{shi2018aster} and CRNN~\cite{shi2016end}. It even surpasses the 3-stage model TPGSR-3 by using only a single stage. 

We also test the inference speed of the three most competitive STISR methods, \ie, TBSRN~($982$ fps), TPGSR~($1,085$ fps) and our TATT model~($960$ fps). TATT has comparable speed with TPGSR and TBSRN, while surpasses them by $2.7\%$ and $3.6\%$ in SR image text recognition by using ASTER as the recognizer. 

To further investigate the performance on spatially deformed text images, we manually pick $804$ rotated and curve-shaped samples from TextZoom test set to evaluate the compared models. Results in Tab.~\ref{table:DistortedTextZoom} indicate that our TATT model obtains the best performance, and the average gap over models like TPGSR and TBSRN becomes larger when encountering spatially deformed text. 

We also visualize the recovery results of both regular samples and spatially-deformed samples of TextZoom in Fig.~\ref{fig:TextZoomVis}. Without TP guidance, TSRN and TBSRN perform far from readable and they are visually unacceptable. With the TP guidance, TPGSR is still unstable in recovering spatially-deformed images. In contrast, our TATT network performs much better in recovering text semantics in samples of all cases compared to all the competitors. With TSC loss, our model further upgrades the visual quality of the samples with better-refined character structure.

\begin{figure}[t]
  \centering
  \includegraphics[width=\linewidth]{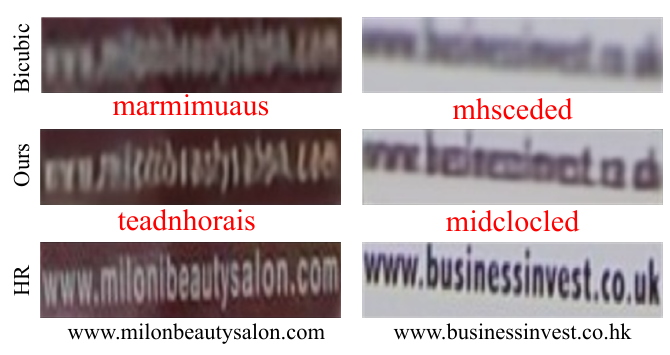}
  \caption{Visualization of the STISR and text recognition results on extremely compressed and blurred text samples.}
  \label{fig:FailureCases}
  \vspace{-0.3cm}
\end{figure}

\textbf{Generalization to recognition dataset.} We evaluate the generalization performance of our TATT network to other real-world text image datasets, including ICDAR15~\cite{karatzas2015icdar}, CUTE80~\cite{risnumawan2014robust} and SVTP~\cite{phan2013recognizing}. These datasets are built for text recognition purpose and contain spatially deformed text image in natural scenes. Since some of the images in these datasets have good quality, we only pick the low-resolution images (\ie, lower than $16 \times 64$) to form our test set with $533$ samples ($391$ from ICDAR15, $3$ from CUTE80 and $139$ from SVTP).  Since the degradation is relatively small, we manually add some degradation on them, including contrast variation, Gaussian noise and Gaussian blurring (see details in \textbf{supplementary file}). We compare with TSRN~\cite{wang2020scene}, TBSRN~\cite{chen2021scene} and TPGSR~\cite{ma2021text} in this test and evaluate the recognition accuracy on the SR results. All models are trained on TextZoom and tested on the picked low-quality images. 

The results are illustrated in Tab.~\ref{table:SynLR_Ablations}, we can see that the proposed TATT network achieves the highest recognition accuracy across all types of degradations. This indicates that our TATT network, though trained on TextZoom, can be well generalized to images in other datasets. The reconstructed high-quality text images by TATT can benefit the downstream tasks such as text recognition.




\section{Conclusion and Discussions}
In this paper, we proposed a Text ATTention network for single text image super-resolution. We leveraged a text prior, which is the semantic information extracted from the text image, to guide the text image reconstruction process. To tackle with the spatially-deformed text recovery, we developed a transformer-based module, called TP Interpreter, to globally correlate the text prior in the semantic domain to the character region in image feature domain. Moreover, we proposed a text structure consistency loss to refine the text structure by imposing structural consistency between the recovered regular and deformed texts. Our model achieved state-of-the-art performance in not only the text super resolution task but the downstream text recognition task.

Though recording state-of-the-art results, the proposed TATT network has limitation on recovering extremely blurry texts, as shown in Fig.~\ref{fig:FailureCases}. In such cases, the strokes of the characters in the text are mixed together, which are difficult to separate. In addition, the computational complexity of our TATT network grows exponentially with the length of the text in the image due to the global attention adopted in our model. It is expected to reduce the computational complexity and improve run-time efficiency of TATT, which will be our future work.

\section{Acknowledgements}
This work is supported by the Hong Kong RGC RIF grant (R5001-18). We thank Dr. Lida Li for the useful discussion on this project.

{\small
\bibliographystyle{ieee_fullname}
\bibliography{camera_ready}
}

\end{document}